\documentclass[letterpaper]{article} 
\usepackage{aaai2026}  

\usepackage{times}  
\usepackage{helvet}  
\usepackage{courier}  
\usepackage[hyphens]{url}  
\usepackage{graphicx} 
\usepackage{natbib}  
\usepackage{caption} 
\usepackage{algorithm}
\usepackage{algorithmic}
\usepackage{algorithm}
\usepackage{algorithmic}
\usepackage{booktabs}
\usepackage{amssymb}
\usepackage{amsmath}
\usepackage{enumitem}
\usepackage{multirow}

\usepackage{newfloat}
\usepackage{listings}
\DeclareCaptionStyle{ruled}{labelfont=normalfont,labelsep=colon,strut=off} 
\floatstyle{ruled}
\newfloat{listing}{tb}{lst}{}
\floatname{listing}{Listing}
\title{\title{BiPrompt: Bilateral Prompt Optimization for Visual and Textual Debiasing in Vision-Language Models}
}

\title{\title{BiPrompt: Bilateral Prompt Optimization for Visual and Textual Debiasing in Vision-Language Models}
}
\author{
    Sunny Gupta\textsuperscript{\rm 1},
    Shounak Das\textsuperscript{\rm 2},
    Amit Sethi\textsuperscript{\rm 2}
}
\affiliations{
    \textsuperscript{\rm 1}Koita Centre for Digital Health, IIT Bombay, Mumbai\\
    \textsuperscript{\rm 2}Department of Electrical Engineering, IIT Bombay, Mumbai\\
    sunnygupta@iitb.ac.in, 21D070068@iitb.ac.in, asethi@iitb.ac.in
}

\usepackage{bibentry}

\begin{document}

\maketitle

\begin{abstract}
Vision language foundation models such as CLIP exhibit impressive zero-shot generalization yet remain vulnerable to spurious correlations across visual and textual modalities. Existing debiasing approaches often address a single modality either visual or textual leading to partial robustness and unstable adaptation under distribution shifts. We propose \textbf{a bilateral prompt optimization framework (BiPrompt)} that simultaneously mitigates non-causal feature reliance in both modalities during test-time adaptation. On the visual side, it employs structured attention-guided erasure to suppress background activations and enforce orthogonal prediction consistency between causal and spurious regions. On the textual side, it introduces balanced prompt normalization, a learnable re-centering mechanism that aligns class embeddings toward an isotropic semantic space. Together, these modules jointly minimize conditional mutual information between spurious cues and predictions, steering the model toward causal, domain invariant reasoning without retraining or domain supervision. Extensive evaluations on real-world and synthetic bias benchmarks demonstrate consistent improvements in both average and worst-group accuracies over prior test-time debiasing methods, establishing a lightweight yet effective path toward trustworthy and causally grounded vision-language adaptation.
\end{abstract}
\section{Introduction}

Vision-Language Models (VLMs), such as CLIP \cite{radford2021learning}, demonstrate remarkable zero-shot generalization by learning from massive image-text datasets. However, their reliability in real-world, out-of-distribution (OOD) settings remains a critical concern.

A fundamental weakness of VLMs is their reliance on decision shortcuts such as spurious correlations and background context rather than true causal features \cite{fan2023improving}. This severely hinders generalization and trustworthiness, leading to unpredictable failures, such as classifying a spider on a beach as a crab. This reliance on spurious features causes incorrect behavior on novel data combinations.

To improve VLM robustness, region-aware methods guide the model's focus by fine-tuning or altering its architecture \cite{sun2024alpha}. However, these approaches are often costly and can harm generalization. A more lightweight alternative is test-time prompt tuning, which adapts a model without modifying its weights.

A prominent example is the Spurious Feature Eraser (SEraser) \cite{ma2025seraser}, which operates on the insight that VLMs possess causal features but are misled by spurious signals. SEraser teaches the model to \textit{ignore} these signals by optimizing a prompt to maximize prediction entropy on auxiliary (spurious) images, forcing the model to rely on causal features.



However, methods like SEraser \cite{ma2025seraser} have two key limitations. First, their reliance on random visual erasure or simplistic segmentation can be unstable, as causal features may be inadvertently removed. Second, they focus exclusively on visual bias, ignoring linguistic biases in the static textual prompts, where strong class-name priors can skew predictions.

To overcome these challenges, we propose \textbf{Balanced-Prompt SEraser (BiPrompt)}, an enhanced test-time adaptation framework that jointly mitigates both visual and textual biases. Our two primary contributions are:
\begin{itemize}
    \item \textbf{Balanced Prompt Normalization}: Learns an isotropic representation for text embeddings at test-time to reduce linguistic bias.
    \item \textbf{Structured Erasure}: Uses attention maps to disentangle causal (foreground) from spurious (background) features, optimizing a prompt to enforce consistency on causal regions while promoting orthogonality to spurious ones.
\end{itemize}

By addressing biases from both modalities, BiPrompt achieves more robust OOD performance without model retraining, contributing to more trustworthy foundation models for real-world applications.
\section{Related Work}
Efforts to improve the out of distribution reliability of Vision Language Models have largely followed two paths. The first involves region aware methods that explicitly guide visual focus, ranging from simple background masking \cite{liang2023mask} to interactive prompting, such as circling the foreground \cite{shtedritski2023redcircle} or using mask contours \cite{yang2023fgvp}. More advanced approaches such as Alpha CLIP \cite{sun2024alpha} extend CLIP with an additional alpha channel, often derived from segmentation models such as SAM, to provide pixel wise information about foreground and background regions. While effective, these methods are typically heavyweight, requiring architectural changes or costly fine tuning on large region text datasets, limiting their out of the box applicability for test time adaptation.
\\

A second, more lightweight approach is test-time prompt tuning, which adapts the model to a new task without modifying its weights. A prominent example is Test-time Prompt Tuning (TPT) \cite{shu2022tpt}, which optimizes a prompt by minimizing the marginal entropy across augmented views of an image, thereby filtering out noisy augmentations based on low confidence. However, this relies on the critical assumption that spurious features will produce low-confidence predictions, which is not always true; a strong spurious feature, like a water background, can erroneously lead to a high-confidence prediction for a land-based object \cite{ma2025seraser}. Other methods leverage language models to refine embeddings, like ROBOSHOT \cite{adila2023roshot}, or debias the model by measuring correlations with biased prompts and applying orthogonal projection \cite{chuang2023debiasing}. Our work builds directly on the insights of Spurious Feature Eraser (SEraser) \cite{ma2025seraser}, which proposed to erase spurious features by maximizing entropy on them. We address these limitations by introducing a structured, attention-guided erasure mechanism and a novel prompt normalization technique to mitigate linguistic bias in both visual and textual modalities.


\begin{figure}[t!]
    \centering

\includegraphics[width=8.4cm,height=11.1cm]{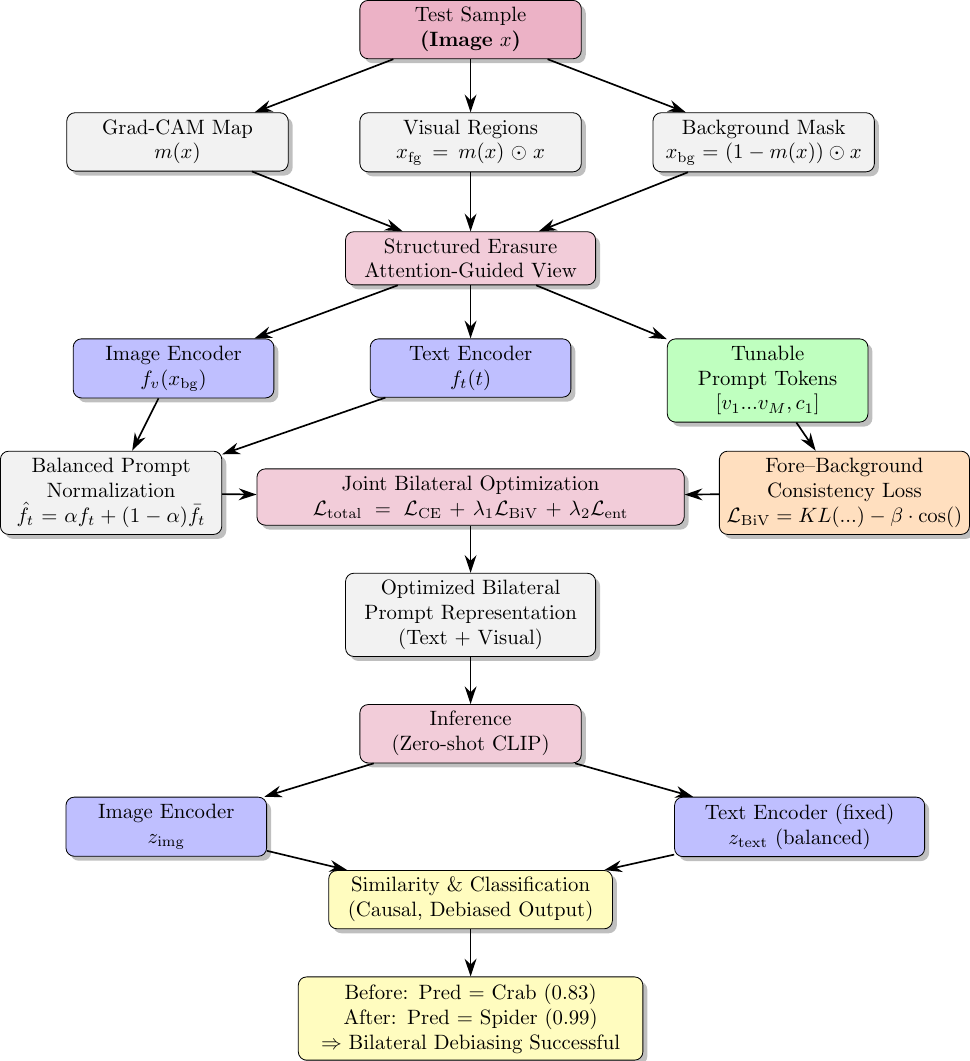}

    \caption{Framework of \textbf{BiPrompt}. Given a test sample, BiPrompt performs bilateral debiasing by jointly optimizing visual and textual representations. Structured attention-guided erasure suppresses spurious visual cues, while balanced prompt normalization aligns textual embeddings. The optimized bilateral prompt is then used for zero-shot inference, yielding causal and robust predictions under distribution shift.}
    \label{fig:biprompt}
\end{figure}
\section{Methodology}

Deep vision--language models (VLMs) such as CLIP have demonstrated remarkable zero-shot generalization across diverse visual domains. 
However, when deployed in \textit{out-of-distribution (OOD)} or \textit{spurious-bias} settings, these models tend to rely on superficial correlations rather than true causal cues. 
For instance, a model might associate ``water'' backgrounds with ``boats'' or ``hospital logos'' with particular pathologies, resulting in degraded reliability under distribution shifts.

Formally, given an image $x$ and a textual class prompt $t_c$, a VLM encodes them into a shared embedding space, where the similarity 
$s(f_v(x), f_t(t_c))$ determines the prediction. 
Under distributional shift, the visual encoder $f_v$ becomes entangled with spurious dimensions $z_s$ (e.g., background, texture), while causal dimensions $z_c$ (object or semantic features) remain underrepresented. 
Our objective is to disentangle and suppress spurious features $z_s$ while preserving causal representations $z_c$, without requiring retraining or access to domain labels.


Given a pretrained CLIP model, predictions are obtained as:
\begin{equation}
p(y|x) = \mathrm{softmax}\!\left(\tau \cdot \mathrm{sim}(f_v(x), f_t(t_y))\right),
\end{equation}
where $\tau$ is a learnable temperature parameter. 
An auxiliary sample $\tilde{x}$ is produced via random erasure, and a KL-divergence regularization aligns their predictions:
\begin{equation}
\mathcal{L}_{\text{SEraser}} = D_{\mathrm{KL}}\big(p(y|x) \,\|\, p(y|\tilde{x})\big).
\end{equation}
To prevent degenerate uniform predictions, SEraser introduces an entropy regularizer:
\begin{equation}
\mathcal{L}_{\text{ent}} = -\frac{1}{C}\sum_{c=1}^{C} p(y=c|x)\log p(y=c|x),
\end{equation}
yielding the total objective:
\begin{equation}
\mathcal{L}_{\text{total}} = 
\mathcal{L}_{\text{CE}} + 
\lambda_1\mathcal{L}_{\text{SEraser}} +
\lambda_2\mathcal{L}_{\text{ent}}.
\end{equation}
This process enforces invariance across erased views, compelling the model to focus on causal regions.

While effective, SEraser suffers from two core issues:
(i) textual prompts remain fixed and thus encode latent linguistic bias, and 
(ii) random erasure can inadvertently mask causal regions, leading to unstable or suboptimal adaptation.

To address these limitations, we introduce \textbf{BiPrompt} a bilateral debiasing framework that simultaneously mitigates visual and textual biases through two complementary mechanisms:
(i) \textit{Balanced Prompt Normalization} for text-space regularization, and
(ii) \textit{Structured Spurious-Region Erasure} for spatially guided visual debiasing.
Together, they enable fine-grained causal alignment without retraining or additional supervision.

\paragraph{(a) Balanced Prompt Normalization.}
Standard prompt embeddings $f_t(t_c)$ often exhibit anisotropy in the textual space, favoring dominant or frequent classes.
To reduce this imbalance, BiPrompt learns a normalized textual embedding:
\begin{equation}
\hat{f}_t(t_c) = \alpha f_t(t_c) + (1-\alpha)\bar{f}_t,
\end{equation}
where $\bar{f}_t = \frac{1}{C}\sum_{c=1}^{C} f_t(t_c)$ is the global semantic centroid, and $\alpha$ is a learnable gating parameter.
This adaptive interpolation encourages isotropic text embeddings, minimizing linguistic dominance and improving stability under domain shift.

\paragraph{(b) Structured Spurious-Region Erasure.}
Instead of random masking, BiPrompt employs \textit{attention-guided erasure}.
Grad-CAM is used to compute soft attention maps $m(x)$ that highlight causal regions. 
We construct complementary foreground and background views:
\begin{equation}
x_{\text{fg}} = m(x)\odot x, \qquad x_{\text{bg}} = (1 - m(x))\odot x.
\end{equation}
Prediction consistency is enforced between $x_{\text{fg}}$ and $x$, while orthogonality is promoted between $x_{\text{bg}}$ and $x$:
\begin{equation}
\mathcal{L}_{\text{BSE}} =
D_{\mathrm{KL}}\!\big(p(y|x_{\text{fg}})\,\|\,p(y|x)\big)
-\beta\,\mathrm{cos}\!\big(p(y|x_{\text{bg}}), p(y|x)\big).
\end{equation}
This structured erasure selectively suppresses spurious activations while preserving causal semantics.

\paragraph{(c) Overall Objective.}
The full test-time objective integrates all components:
\begin{equation}
\mathcal{L}_{\text{total}} =
\mathcal{L}_{\text{CE}} +
\lambda_1\mathcal{L}_{\text{BSE}} +
\lambda_2\mathcal{L}_{\text{ent}}.
\end{equation}
Only a few lightweight parametersthe gating $\alpha$ and normalization weightsare updated during adaptation, keeping the optimization efficient and memory-friendly.

\subsection{Optimization and Inference}

During inference, BiPrompt performs the following lightweight test-time adaptation steps:
\begin{enumerate}[nosep,leftmargin=*]
    \item Compute Grad-CAM attention maps to obtain foreground and background views $(x_{\text{fg}}, x_{\text{bg}})$.
    \item Extract visual features $f_v(x)$, $f_v(x_{\text{fg}})$, and $f_v(x_{\text{bg}})$, along with normalized text embeddings $\hat{f}_t(t_c)$.
    \item Perform one or few gradient updates to minimize $\mathcal{L}_{\text{total}}$.
    \item Compute final predictions using similarity $s(f_v(x), \hat{f}_t(t_c))$.
\end{enumerate}
This adaptive process aligns visual and textual embeddings across causal regions, improving generalization to unseen domains. BiPrompt implicitly minimizes the conditional mutual information between spurious features $z_s$ and predictions $y$:
\begin{equation}
I(z_s; y|z_c) \approx 0.
\end{equation}
The structured erasure term enforces conditional independence in the visual space, while balanced prompt normalization reduces anisotropy in the textual space. 
Together, these mechanisms yield a causally disentangled representation that enhances robustness and reliability under distribution shift.

\begin{table*}[ht]
\centering
\caption{Zero-shot classification performance on real-world OOD datasets (Tiny-ImageNet, CUB-200, ImageNet-A). Top-1 accuracy (\%) and improvement over Vanilla are shown.}
\label{tab:realworld}
\begin{tabular}{lcccccccc}
\toprule
Dataset & Vanilla & TPT & RoSHOT & $\alpha$-CLIP & Patches & Images & Blocks & \textbf{BiPrompt (Ours)} \\
\midrule
Tiny-ImageNet & 23.2 & 29.6 & 49.2 & 76.0 & 42.4 & 41.2 & 42.8 & \textbf{44.1} ($\blacktriangle$20.9) \\
CUB-200       & 12.1 & 8.7  & 25.5 & 44.3 & 26.2 & 24.2 & 28.9 & \textbf{31.0} ($\blacktriangle$18.9) \\
ImageNet-A    & 42.1 & 49.7 & 38.9 & 51.5 & 47.4 & 45.4 & 49.7 & \textbf{52.2} ($\blacktriangle$10.1) \\
\midrule
\textbf{Average} & 25.8 & 29.3 & 37.9 & 57.3 & 38.7 & 36.9 & 40.5 & \textbf{42.4} ($\blacktriangle$16.6) \\
\bottomrule
\end{tabular}
\end{table*}

%

%


\begin{table*}[t]
\centering
\caption{
Zero-shot classification performance on simulated spurious-bias scenarios.
Average accuracy (AVG.) and worst-group accuracy (W.G.) are reported across Waterbirds, CamelDeer, and SpiderCrab datasets.
The symbols $\uparrow$ and $\downarrow$ denote performance gain or drop relative to the Vanilla baseline for BiPrompt.
}
\label{tab:biPrompt_results}
\setlength{\tabcolsep}{5.5pt}
\renewcommand{\arraystretch}{1.1}
\begin{tabular}{lccccccccc}
\toprule
\textbf{Dataset} & \textbf{Metric} & \textbf{Van.} & \textbf{MASK} & \textbf{TPT} & \textbf{RoSHOT} & \textbf{$\alpha$-CLIP} & \textbf{SEraser} & \textbf{BiPrompt} \\
\midrule
\multirow{2}{*}{Waterbirds} 
 & AVG. & 67.7 & 72.0 & 66.9 & 68.9 & 67.6 & 78.2 & \textbf{79.9 ($\uparrow$ 12.2)} \\
 & W.G. & 40.0 & 51.5 & 34.4 & 52.3 & 43.2 & 65.3 & \textbf{66.6 ($\uparrow$ 26.5)} \\
\midrule
\multirow{2}{*}{CamelDeer} 
 & AVG. & 83.2 & 93.6 & 77.7 & 80.4 & 92.0 & 95.7 & \textbf{97.2 ($\uparrow$ 14.0)} \\
 & W.G. & 66.4 & 87.2 & 55.3 & 60.8 & 84.4 & 91.6 & \textbf{92.8 ($\uparrow$ 26.4)} \\
\midrule
\multirow{2}{*}{SpiderCrab} 
 & AVG. & 66.0 & 91.4 & 83.5 & 73.0 & 86.2 & 95.3 & \textbf{97.4 ($\uparrow$ 31.4)} \\
 & W.G. & 42.0 & 90.4 & 72.5 & 50.4 & 86.0 & 94.7 & \textbf{95.4 ($\uparrow$ 53.4)} \\
\midrule
\multirow{2}{*}{Avg. (3 sets)} 
 & AVG. & 72.3 & 85.7 & 76.0 & 74.1 & 81.9 & 89.8 & \textbf{91.3 ($\uparrow$ 19.0)} \\
 & W.G. & 49.5 & 76.4 & 54.1 & 54.5 & 71.2 & 83.7 & \textbf{85.0 ($\uparrow$ 35.5)} \\
\bottomrule
\end{tabular}
\end{table*}


\begin{table}[t]
\centering
\caption{
Zero-shot classification performance of different VLFMs on the Waterbirds dataset.
Average accuracy (AVG.) and worst-group accuracy (W.G.) are reported.
The symbol $\blacktriangle$ indicates performance gain over the Vanilla baseline.
}
\label{tab:vlfm}
\setlength{\tabcolsep}{2.8pt}
\renewcommand{\arraystretch}{1.05}
\begin{tabular}{lcccc}
\toprule
\textbf{Model} & \textbf{Van.} & \textbf{MASK} & \textbf{SEraser} & \textbf{BiPrompt} \\
\midrule
\textbf{CLIP-L14} (AVG.) & 83.7 & 85.5 & 87.8 & \textbf{88.4} ($\blacktriangle$4.7) \\
\quad W.G. & 32.9 & 40.8 & 58.9 & \textbf{60.1} ($\blacktriangle$27.2) \\[3pt]
\textbf{BLIP-2} (AVG.) & 57.7 & 54.4 & 55.6 & \textbf{56.3} ($\blacktriangle$-1.4) \\
\quad W.G. & 28.2 & 35.1 & 34.7 & \textbf{35.5} ($\blacktriangle$7.3) \\
\midrule
\textbf{Avg. (2 sets)} (AVG.) & 70.7 & 69.9 & 71.7 & \textbf{72.4} ($\blacktriangle$1.7) \\
\quad W.G. & 30.5 & 37.9 & 46.8 & \textbf{47.8} ($\blacktriangle$17.3) \\
\bottomrule
\end{tabular}
\end{table}

\section{Results}

We evaluate the effectiveness of our proposed method, BiPrompt, against several baselines on a diverse set of benchmarks. We first describe the experimental setup, then present the main results on both real-world OOD datasets and simulated spurious-bias scenarios, followed by an ablation study on model generality.

\subsection{Experimental Setup}

\paragraph{Datasets.} 
We evaluate BiPrompt across two experimental settings. 
For \textit{real-world out-of-distribution (OOD)} data, we use Tiny-ImageNet~\cite{le2015tiny}, CUB-200~\cite{wah2011cub}, and ImageNet-A, which capture naturally occurring domain shifts. 
For \textit{simulated spurious-bias} data, we adopt the Waterbirds benchmark~\cite{koh2021wilds} and two datasets generated by the S2E protocolCamelDeer and SpiderCrab~\cite{ma2025seraser}. 
These datasets explicitly model backgroundobject correlations, allowing us to assess BiPrompt's ability to mitigate shortcut learning and maintain causal generalization under controlled bias.

\paragraph{Baselines and Backbone.} 
We compare BiPrompt with several representative methods, including Vanilla CLIP, TPT~\cite{shu2022tpt}, RoSHOT~\cite{adila2023roshot}, $\alpha$-CLIP, and SEraser~\cite{ma2025seraser}. 
All methods, except $\alpha$-CLIP which uses its own encoder, are built on the pre-trained CLIP ViT-B/32 backbone~\cite{radford2021learning}. 
For SEraser, we adopt its \textit{Patches} variant using four corner patches from an $8\times8$ gridand the \textit{Images} variant that leverages OOD reference samples. 
In simulated bias experiments, all test-time adaptation methods (including BiPrompt) employ SAM~\cite{kirillov2023segment} to isolate and erase background regions, ensuring fair and consistent evaluation across frameworks.

\subsection{Results on Real-World OOD Scenarios}

As shown in Table~\ref{tab:realworld}, our proposed BiPrompt consistently outperforms all baselines on average across the three real-world datasets. Notably, BiPrompt achieves a Top-1 accuracy of \textbf{42.4\%}, surpassing the strongest original SEraser strategy (Blocks) by 1.9\%. This demonstrates the clear benefit of our dual-pronged approach, which addresses both visual and linguistic biases.

On CUB-200 and ImageNet-A, BiPrompt shows significant gains of 2.1\% and 2.5\%, respectively, over the next-best SEraser variant. While $\alpha$-CLIP achieves an outstanding performance on Tiny-ImageNet, this is expected as its checkpoint was retrained on ImageNet \cite{ma2025seraser}. On the other hand, less-biased benchmarks, such as BiPrompt, prove to be the most robust method.

\subsection{Results on Simulated Spurious-Bias Scenarios}

Table~\ref{tab:biPrompt_results} presents the results on datasets explicitly designed to measure robustness against spurious correlations, with a focus on average (AVG) and worst-group (W.G.) accuracy. In these challenging scenarios, BiPrompt demonstrates a substantial improvement over all other methods.

Across all three datasets, BiPrompt achieves the highest worst-group accuracy, validating its superior ability to mitigate decision shortcuts. This is particularly evident on the Waterbirds benchmark, where BiPrompt improves the W.G. performance significantly over the Vanilla baseline.

Our method significantly outperforms the original SEraser, which, despite its strong performance, is surpassed by BiPrompt's structured erasure and prompt normalization. In contrast, other methods, such as TPT, perform suboptimally. In contrast, other methods such as TPT assume that high-confidence views capture invariant features, but this often fails when strong spurious cues, such as a desert background, lead to confident yet incorrect predictions. RoSHOT exhibits similarly inconsistent behavior, whereas the consistent and notable gains, particularly in worst-group accuracy, demonstrate the robustness and effectiveness of our method.

\subsection{Effectiveness Across Different Model Architectures}

To evaluate the flexibility and generality of our approach, we tested BiPrompt on other widely-used VLM architectures, specifically \textbf{CLIP ViT-L-14} and \textbf{BLIP-2}. We conducted this evaluation on the challenging Waterbirds benchmark \cite{ma2025seraser}, with results shown in Table~\ref{tab:vlfm}. Our method achieves promising performance on both models, consistently outperforming the baseline.
\section{Conclusion}

In this work, we tackled the challenge of unreliable generalization in vision--language models (VLMs) that often depend on spurious correlations when faced with out-of-distribution (OOD) data. 
While existing test-time methods primarily target visual bias, they often overlook the interplay between visual and linguistic factors. 
We introduced \textbf{BiPrompt}, a bilateral test-time adaptation framework that jointly mitigates both forms of bias through \textit{Balanced Prompt Normalization} for text-space isotropy and \textit{Structured Erasure} for attention-guided visual debiasing. 
Extensive experiments demonstrate that BiPrompt achieves consistent zero-shot gains, particularly on worst-group accuracy benchmarks. 
By enabling cross-modal bias correction without retraining, BiPrompt offers a simple yet effective step toward building more reliable and causally grounded foundation models.
\bibliography{aaai2026}


\end{document}